\documentclass[a4paper]{article}
\usepackage{cite} 
\usepackage[a4paper]{geometry}\usepackage{amsthm}
\usepackage[T1]{fontenc}
\usepackage[utf8]{inputenc}  
\usepackage{gensymb}
\usepackage{amsfonts}
\usepackage{graphicx}
\usepackage{url}
\usepackage{booktabs}
\usepackage{siunitx}
\usepackage{units}
\usepackage[]{algorithm}
\usepackage{algpseudocode}
\usepackage{listings}
\input{atomone.pygstyle}
\input{default-pyg-prefix.pygstyle}
\usepackage[frozencache]{minted} 
\usepackage{multicol}
\usepackage[ntheorem]{empheq}

\usepackage{fourier}\usepackage[scale=0.85]{tgheros}\DeclareSymbolFont{symbols2}     {OMS}{cmsy}{m}{n}\DeclareSymbolFontAlphabet{\mathcal}{symbols2}

\usepackage{calc} 
\usepackage{color}
\usepackage{pgfplots}
\pgfplotsset{compat=newest}
\pgfplotsset{plot coordinates/math parser=false}
\pgfplotsset{legend style={draw=white}}
\usepgfplotslibrary{groupplots}
\usetikzlibrary{plotmarks}
\usetikzlibrary{positioning}
\usetikzlibrary{shapes,arrows}
\usetikzlibrary{through}
\tikzset{block/.style={draw, rectangle, line width=2pt,minimum height=2em, minimum width=3em, outer sep=0pt}}
\usepgfplotslibrary{patchplots}
\definecolor{red}{rgb}{1,0,0}
\definecolor{gray}{rgb}{.55,.5,.5}
\definecolor{green}{rgb}{0,.9,0}
\definecolor{blue}{rgb}{0,0,0.55}
\definecolor{yellow}{rgb}{.5,.5,0}
\definecolor{darkred}{rgb}{0.5,0,0}
\definecolor{darkgreen}{rgb}{0,0.6,0}
\definecolor{darkblue}{rgb}{0,0,0.6}
\definecolor{orange}{rgb}{1,0.647058823529412,0}
\definecolor{cyan}{rgb}{0,1,1}

\definecolor{linecolor1}{rgb}{0.1399999, 0.1399999, 0.4}
\definecolor{linecolor2}{rgb}{1.0, 0.7075, 0.35}
\definecolor{linecolor3}{rgb}{0.414999, 1.0, 1.0}
\definecolor{linecolor4}{rgb}{0.6, 0.21, 0.534999}

\pgfplotsset{
backgroundstyle/.style={
axis background style={fill=gray!6}
}}

\usepgfplotslibrary{external}
\tikzsetexternalprefix{build/}

\newcommand{\recmt}[1]{}

\newcommand{\rcmt}[1]{}

\DeclareMathOperator*{\argmin}{arg\,min}
\DeclareMathOperator{\sign}{sign}
\renewcommand{\skew}[1]{\left<#1\right>}
\newcommand{\norm}[1]{\begin{Vmatrix}#1\end{Vmatrix}}
\newcommand{\normf}[1]{\begin{Vmatrix}#1\end{Vmatrix}_F}
\newcommand{\normt}[1]{\begin{Vmatrix}#1\end{Vmatrix}_2}

\newcommand{\RTFS}{{R_{\scriptscriptstyle TF}^{S}}}
\newcommand{\RSTF}{{R_{S}^{\scriptscriptstyle TF}}}
\newcommand{\RRBTF}{{R_{\scriptscriptstyle RB}^{\scriptscriptstyle TF}}}
\newcommand{\RTFRB}{{R_{\scriptscriptstyle TF}^{\scriptscriptstyle RB}}}

\newcommand{\T}{^{\hspace{-0.1mm}\scriptscriptstyle \mathsf{T}}\hspace{-0.2mm}}

\newcommand{\inspace}[1]{\in \mathbb{R}^{#1}}

\renewcommand{\vec}{\operatorname{vec}}

\newcommand{\bmatrixx}[1]{\begin{bmatrix}#1\end{bmatrix}}

\newcommand{\softwarecite}[1]{\cite{#1}} 

\usepackage{microtype}

\usepackage[capitalise]{cleveref} 
\crefname{section}{Sec.}{Sections}
\Crefname{section}{Section}{Sections}
\crefname{chapter}{Chap.}{Chapters}
\Crefname{chapter}{Chapter}{Chapters}
\crefname{proposition}{Proposition}{Propositions}
\Crefname{proposition}{Proposition}{Propositions}
\crefname{corollary}{Corollary}{Corollaries}
\Crefname{corollary}{Corollary}{Corollaries}
\crefname{lemma}{Lemma}{Lemmas}
\Crefname{lemma}{Lemma}{Lemmas}
\crefname{objective}{Objective}{Objectives}
\Crefname{objective}{Objective}{Objectives}
\crefname{algorithm}{Algorithm}{Algorithms}
\Crefname{algorithm}{Algorithm}{Algorithms}

\begin{document}
\newtheorem{proposition}{Proposition}
\newtheorem{corollary}{Corollary}
\newtheorem{lemma}{Lemma}
\newtheorem{theorem}{Theorem}
\newtheorem{remark}{Remark}
\newtheorem{definition}{Definition}
\newlength\figureheight
\newlength\figurewidth
\setlength{\figurewidth}{0.4\textwidth}
\setlength{\figureheight }{4cm }

	\title{On the Calibration of Force/Torque Sensors in Robotics}
	\author{Fredrik Bagge Carlson\\ \texttt{fredrikb@control.lth.se}}
%

\maketitle

\begin{abstract}
	We present and analyze methods for the kinematic and kinetostatic calibration of, typically, wrist mounted force/torque sensors in robotics. The algorithms are based on matrix factorization and require no special equipment. The only requirement is the ability to reorient the sensor and to measure its orientation in a fixed coordinate system, such as through the forward kinematics of a robot manipulator, or using an external tracking system. We present methods to find the rotation matrix between the coordinate system of the sensor and that of the tool flange, the mass held by the force sensor at rest, the vector to the center of gravity of this mass and the gravitational acceleration vector.
\end{abstract}

\section{Introduction}

A 6\ DOF force/torque sensor is a device capable of measuring the complete wrench of forces and torques applied to the sensor. They are commonly mounted on the tool flange of a manipulator to endow it with force/torque sensing capabilities, useful for, e.g., accurate control in contact situations.

In order to make use of a force/torque sensor, the rotation matrix $\RTFS$ between the tool flange and the sensor coordinate systems, the mass $m$ held by the force sensor at rest, and the translational vector $r \inspace{3}$ from the sensor origin to the center of mass are required. Methods from the literature typically involve fixing the force/torque sensor in a jig and applying known forces/torques to the sensor \cite{song2007novel,chen2015design}. In the following, we will develop and analyze calibration methods that only require movement of the sensor attached to the tool flange in free air, making them very simple to use.

This work was partly presented in the thesis \cite{bagge2018}, and is extending the previously presented material for cases where the gravity vector is unknown.

\subsection{Notation}
The notation in this work is summarized in \cref{tab:variables}. The coordinate frame of the tool-flange is denoted $\mathcal{TF}$. This is the mechanical interface between the robot and the payload or tool, in our case the force/torque sensor. The robot base frame $\mathcal{RB}$ is the base of the forward-kinematics function of a manipulator, but could also be, e.g., the frame of an external optical tracking system that measures the location of the tool frame in the case of a flying robot etc. A sensor delivers measurements in the sensor frame $\mathcal{S}$. A vector $x$ given in coordinate-frame $\mathcal{RB}$ is denoted $x_{RB}$, and is rotated to the frame $\mathcal{TF}$ by application of a rotation matrix according to $x_{TF} = \RTFRB x_{RB}$

The matrix $\skew{s} \in so$ is formed by the elements of a vector $s$ and has the skew-symmetric property $\skew{s} + \skew{s}\T = 0$ \cite{murray1994mathematical}.
\begin{table}[htb]
    \centering
    \caption{Definition and description of coordinate frames, variables and notation.}
    \begin{tabular}{@{}lll@{}}
        \toprule
        $\mathcal{RB}$ &~&Robot base frame.\\
        $\mathcal{TF}$ &~&Tool-flange frame, attached to the TCP.\\
        $\mathcal{S}$  &~&Sensor frame.\\
        $\tau $ & $\inspace{n}$         & Torque vector.           		        \\
        $\mathfrak{f}$ & $\inspace{6}$  & External force/torque wrench.        	\\
		$m$ & $\inspace{}$  			& Mass held by the sensor.        	\\
		$g$ & $\inspace{3}$  			& Gravitational acceleration vector.  	\\
        $R_A^B $ & $\in SO(3)$          & Rotation matrix from $\mathcal{B}$ to $\mathcal{A}$.\\
        $\skew{s}$   & $\in so(3)$      & Skew-symmetric matrix with parameters $s \inspace{3}$.\\
        $\nabla_x f$ & ~                & Gradient of $f$ with respect to $x$.	\\
        $\hat{x}$    & ~                & Estimate of variable $x$.				\\
        $x_{i:j}$    & ~                & Elements $i, i+1, ..., j$ of $x$.		\\
		$A^\dagger$  & ~                & The pseudo-inverse of $A$: $(A\T A) ^{-1} A\T$.\\
        \bottomrule
    \end{tabular}
    \label{tab:variables}
\end{table}

\pagebreak
\section{Calibration with known gravity vector} \label{sec:forcecalib_method}
We start with a simplified case, where the direction, but not the magnitude, of the gravity vector is known. This case is common in practice, since the gravity vector commonly coincides with the negative $z$-axis of a robot mounted upright on the floor. This case also demonstrates how to estimate the vector to the center of gravity of the mass held by the force sensor, given the rotation matrix and the gravitational vector.

The relevant force and torque equations are given by
\begin{align}
     f_S &= \RSTF \RTFRB (mg_{RB})\label{eq:forceeq} \\
     \tau_S &= \RSTF\skew{r}\RTFRB (mg_{RB})\label{eq:torqueeq}
\end{align}
where $g$ is the gravity vector given in the robot base-frame with $\norm{g} \approx \SI{9.8}{m/s^2}$ and $f,\tau$ are the force and torque measurements, respectively, such that $\mathfrak{f} = [f\T \; \tau\T]\T$. At first glance, this is a hard problem to solve.
The equation for the force relation does not appear to allow us to solve for both $m$ and $\RSTF$, the constraint $R \in SO(3)$ is difficult to handle, and the equation for the torque contains the nonlinear term $\RSTF\skew{r}$. Fortunately, however, the problem can be separated into two steps, and the constraint $\RSTF \in SO(3)$ will allow us to distinguish $\RSTF$ from $m$.

A naive approach to the stated calibration problem is to formulate an optimization problem where $R$ is parameterized using, e.g., Euler angles. A benefit of this approach is its automatic and implicit handling of the constraint $R \in SO(3)$. One then minimizes the residuals of \labelcref{eq:forceeq} and \labelcref{eq:torqueeq} with respect to all parameters using a local optimization method. This approach is, however, prone to convergence to local optima and is hard to conduct in practice.

Instead, we start by noting that multiplying a matrix with a scalar only affects its singular values, but not its singular vectors, $mR = U(mS)V\T$. Thus, if we solve a convex relaxation to the problem and estimate the product $mR$, we can recover $R$ by projecting $mR$ onto $SO(3)$ using the procedure in~\cref{sec:orthonormal}. Given $R$ we can easily recover $m$. \Cref{eq:forceeq} is linear in $mR$ and the minimization step can readily be conducted using standard linear least-squares. To facilitate this estimation, we write \labelcref{eq:forceeq} on the equivalent form
\begin{equation}
    (f_S\T \otimes I_3)\vec(mR) = \RTFRB g
\end{equation}
where $\vec(mR) \inspace{9}$ is a vector of parameter to be estimated.

Once $\RSTF$ and $m$ are estimated using measured forces only, we can estimate $r$ using the torque relation by noting that
\begin{align}
    \tau_S &= \RSTF\skew{r}\RTFRB (mg)\\
    \RTFS\tau_S &= \skew{\RTFRB (mg)} r
\end{align}
where the second equation is linear in the unknown parameter-vector $r$.

When solving a relaxed problem, there is in general no guarantee that a good solution to the original problem will be found. To verify that relaxing the problem does not introduce any numerical issues or problems in the presence of noise, etc., we introduce a second algorithm for finding $\RSTF$. If the mass $m$ is known in advance, the problem can be reformulated using the Cayley transform~\cite{tsiotras1997higher} and a technique similar to the attitude estimation algorithm found in~\cite{mortari2007olae} can be used to solve for the rotation matrix, without constraints or relaxations.

The Cayley transform of a matrix $R \in SO(3)$ is given by
\begin{equation}
    R = (I+\Sigma)^{-1} (I-\Sigma) = (I-\Sigma)(I+\Sigma)^{-1}
\end{equation}
 where $\Sigma = \skew{s}$ is a skew-symmetric matrix of the Cayley-Gibbs-Rodrigues parameters $s \inspace{3}$ \cite{tsiotras1997higher}. Applying the Cayley transform to the force relation yields
\begin{align}
    f_S &= \RSTF \RTFRB (mg)\\
    f_S &= (I+\Sigma)^{-1} (I-\Sigma) \RTFRB (mg)\\
    (I+\Sigma)f_S &= (I-\Sigma) \RTFRB (mg)\\
    \Sigma \big(f_S + \RTFRB (mg)\big)  &= -\big(f_S - \RTFRB (mg)\big)\\
    \Big[ \Sigma = \skew{s}, &\quad \skew{a}b = \skew{b}a \Big]\nonumber\\
    \skew{f_S + \RTFRB (mg)}s  &= -\big(f_S - \RTFRB (mg)\big)
\end{align}
which is a linear equation in the parameters $s$ that can be solved using the standard least-squares procedure.\footnote{An open-source implementation that solves the problem, as well as code to reproduce the figures in the numerical evaluation, is provided in \softwarecite{Robotlib}.} The least-squares solution to this problem was, however, found during experiments to be very sensitive to measurement noise in $f_S$. This is due to the fact that $f_S$ appears not only in the dependent variable on the right-hand side, but also in the regressor $\skew{f_S + \RTFRB (mg)}$. This is thus an \emph{errors-in-variables} problem for which the solution is given by the \emph{total least-squares} procedure \cite{golub2012matrix}, which we make use of in the following evaluation.

\section{Calibration with unknown gravity vector}
When the gravity vector $g$ is unknown or uncertain, the estimation problem is more difficult.
The equation
\begin{equation}\label{eq:forceeq2}
    f_S = \RSTF \RTFRB (mg)
\end{equation}
is not linear in both $g$ and $\RSTF$ simultaneously.
A reformulation into a linear system according to
\begin{equation}
    \RTFS f_S - \RTFRB g = 0
\end{equation}
where we have included the unknown mass $m$ into $g$, appears to allow us to solve for the parameters in both $g$ and $\RTFS$. Unfortunately, the constraint $\RTFS \in SO(3)$ now becomes prohibitively difficult to handle, as the relaxed problem has the trivial solution $\RTFRB = g = 0$. In the following, we present a series of solution methods that solve the problem equally well when signal-to-noise ratio (SNR) is high. Two methods are based on eigendecompositions, one on polynomial optimization through homotopy continuation and one based on iterating between solving two linear least-squares problems, which will turn out to be yield better results when the SNR approaches 1.

\subsection{Solution by eigendecomposition}

The optimal $\RTFS$ is available as an eigenvector to a particular matrix, as detailed in the following theorem.
\begin{theorem}\label{theorem1}
    Introduce the linear maps
    \begin{align}
        \mathcal{F} &: g \rightarrow (F\T F)^{-1}F\T D g\\
        \mathcal{D} &: r \rightarrow (D\T D)^{-1}D\T F r\\
        \mathcal{K} &:\mathcal{F} \circ \mathcal{D} = (F\T F)^{-1}F\T D (D\T D)^{-1}D\T F = F^\dagger D D^\dagger F
    \end{align}
    where
    \begin{equation}
        D = \bmatrixx{\RTFRB_1 \\ \RTFRB_2 \\ \vdots} \qquad
        F = \bmatrixx{f_1\T \otimes I_3 \\ f_2\T \otimes I_3 \\ \vdots}
    \end{equation}
    where $D$ and $F$ both have full column rank.\footnotetext{This is easily accomplished by sufficient reorientation of the sensor.}

    The matrix $\RTFS$ is then given by the eigenvector $r^*$ of the matrix $\mathcal{K}$ associated with the eigenvalue 1.
    \begin{proof}
        We will interchangebly refer to both the matrix $R$ and the vector $r$, where $R = \operatorname{reshape}(r, 3, 3)$ or $r = \vec(R)$.

        We can write the problem of finding $r = \vec(\RTFS)$ given $g$ as
        \begin{equation}
            \bar r = \argmin_r \normt{Fr - Dg}
        \end{equation}
        with the closed-form solution $\bar r = (F\T F)^{-1}F\T D g = \mathcal{F}g$.
        We can similarly write the problem of finding $g$ given $r$ as
        \begin{equation}
            \argmin_g \normt{Fr - Dg}
        \end{equation}
        with the closed-form solution
        \begin{equation}\label{eq:g}
            g = (D\T D)^{-1}D\T F r = \mathcal{D}r
        \end{equation}

        Given an initial guess $r_0$, the composition of the two stated optimization problems can now be written
        \begin{equation}
            r_1 = \mathcal{F} \circ \mathcal{D} \; r_0 = \mathcal{K} \; r_0
        \end{equation}

        In the noise-free case, we can immediately conclude that the true solution $R^*$ is a fixed point of the operator $\mathcal{K}$ due to the uniqueness-properties of the least-squares solution.
        This implies that the true solution $R^*$ is an eigenvector of the linear operator $\mathcal{K}$ with eigenvalue 1.
        $R^*$ can thus be found by eigendecomposition of $\mathcal{K}$. This eigenvalue must be unique, as otherwise there would be several solutions to the intermediate optimization problems, implying that the matrices $F$ and $D$ are rank deficient.
        The correct sign of $r$ is determined such that $\sign\det(R) = 1$ and the scaling such that $\norm{R} = 1$.

    \end{proof}
\end{theorem}
In the case when the data $f_S$ is corrupted by measurement noise, we follow the calculation of the eigenvector of $\mathcal{K}$ with a projection onto $SO(3)$. When $R$ is found, $g$ is calculated from \labelcref{eq:g}.

Code to calculate these solutions are given in \cref{sec:implementations}.

\subsection{Nullspace method}
Since both $R = \RTFS$ and $g$ are constant, one can form the difference between two measurements to get rid of the variable $g$:
\begin{align}
    A_1\T R f_1 &= A_2\T R f_2 &&= g \\
    A_1\T R f_1 &- A_2\T R f_2 &&= 0\\
    A_2A_1\T R f_1 &-  R f_2 &&= 0\\
    A_2A_1\T (f_1\T \otimes I_3)\vec(R) &-  (f_2\T \otimes I_3)\vec(R) &&= 0\\
	\big(A_2A_1\T (f_1\T \otimes I_3) &-  (f_2\T \otimes I_3) \big)\vec(R) &&= 0
\end{align}
where $A$ denotes $\RRBTF$. This is now an equation system in the variable $r = \vec(R)$ only. It is now clear that $r$ must lie in the nullspace of the matrix on the left-hand side. By stacking several of these equations systems, the nullspace is reduced to one dimension, the solution we are looking for.

This method only works if we can guarantee that the nullspace will be one dimensional given enough data. $\RTFS$ is the unique element in $SO(3)$ that solves equation \labelcref{eq:forceeq2}, but without this constraint, the equation system has a six-dimensional solution space. However, given sufficient amount of data which is sufficiently diverse, the solutions-space is reduced to one dimension. Once $R$ is estimated, finding $g$ amounts to solving a linear least-squares problem.
The nullspace of a matrix $M$ is easily found using the singular-value decomposition $USV\T = M$ as the right singular vectors\footnote{I.e., columns of $V$ (or rows of $V\T$).} corresponding to singular values of 0. When the data is corrupted by noise, these singular values will be non-zero but small.
An implementation of this algorithm is provided in \cref{sec:implementations}.

\subsection{Polynomial Optimization}

\Cref{eq:forceeq2} is a polynomial equation system (quadratic), and the constraint $R \in SO(3)$ can be relaxed slightly and written as the polynomial system $R\T R = I$, which allows matrices with determinant $\pm 1$. To solve for all parameters, respecting the relaxed constraint, we form the Lagrangian
\begin{equation}
    L = \sum \big( f_S - \RSTF A g\big)^2 + \lambda\T \vec(R\T R - I)
\end{equation}
and use homotopy continuation to find all solutions to $\nabla L = 0$, where $\nabla L$ denotes the gradient of $L$ with respect to all parameters, including the Lagrange multipliers $\lambda \inspace{9}$.
Trying to solve this system will result in a large number of solution paths ($2^{18}$), which is feasible and takes less than 10 minutes.
The desired solution is easily identified as the solution yielding the lowest value of $L$ while having $\det(R) = 1$.
The solution time can be reduced slightly by reducing the number of Lagrange multipliers to 6, resulting in a problem with $2^{15}$ solution paths, which can be solved in the order of \SI{10}{s}.

Our implementation made use of \softwarecite{HomotopyContinuation}.

\subsection{Iterative Linear Least-Squares}
We noted earlier that solving for $\RTFS$ given $g$ is possible using standard techniques. We can further note that given $\RTFS$, \labelcref{eq:forceeq2} is linear in $g$. This opens up for the possibility of providing an initial guess of $g$, solving for $\RTFS$ and using the resulting rotation matrix after projection onto $SO(3)$ to find an updated guess for $g$.
This approach will resemble that of power iteration, an algorithm used to find the largest eigenvector of a matrix $K$ by repeatedly multiplying a vector by $K$, with intermediate renormalization. The difference lies in the renormalization applied, which in our case is a projection onto $SO(3)$ as opposed to a simple rescaling. \Cref{theorem1} established the true solution as the largest eigenvector of the composition of the two optimization problems, making it believable that this algorithm will converge. Convergence of the algorithm, provided in \cref{alg:iterative}, is verified in \cref{sec:evaluation} using noisy data and arbitrary initial guesses of $g$.

\begin{algorithm}
    \caption{Pseudo-code for joint estimation of rotation matrix and gravity vector using the iterative least-squares procedure. An open-source implementation is provided in \softwarecite{Robotlib}.}
    \label{alg:iterative}
    \begin{algorithmic}
        \State $g \gets g_0$ \Comment{Initialize $g$ (a vector of zeros works well).}
        \Repeat
            \State $\bar R \gets \argmin_{R} \sum_{i\in 1:N} \normf{Rf_{S_i} - \RTFRB_i g}^2$ \Comment{Solve the problem of \cref{sec:forcecalib_method}}
            \State $\RTFS \gets \operatorname{project}(\bar R)$ \Comment{Project onto $SO(3)$}
            \State $g \gets \argmin_{g} \sum_{i\in 1:N} \normf{\RTFS f_{S_i} - \RTFRB_i g}^2$ \Comment{Solve for a new $g$ using linear least-squares.}
        \Until{convergence}
    \end{algorithmic}
\end{algorithm}

\section{Numerical evaluation} \label{sec:evaluation}

\subsection{Known gravity vector}

\begin{figure}[b]
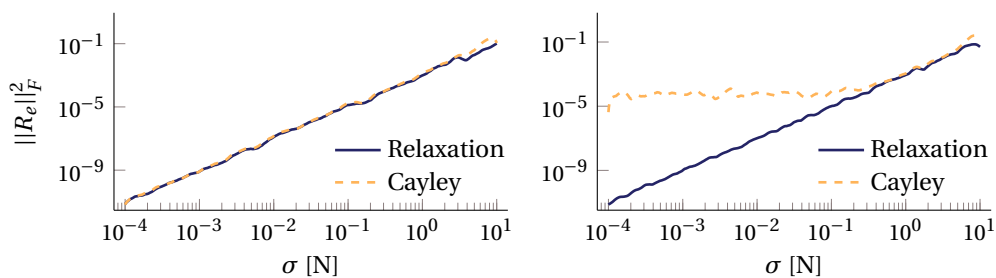

    \centering
    \setlength{\figurewidth}{0.46\linewidth}
    \setlength{\figureheight }{4.2cm}
    \input{figs/calibforce.tex}
    \input{figs/calibforce2.tex} 
    \caption{The error in the estimated rotation matrix is shown as a function of the added measurement noise for two force-calibration methods, relaxation based and Cayley-transform based. On the left, the relaxation-based method was started with an initial mass estimate $m_0 = 2m$ whereas the Cayley-transform based method was given the correct mass. On the right, both algorithms were given $m_0 = 1.1m$ }
    \label{fig:calibforce}
\end{figure}

The two algorithms, the relaxation-based and the Cayley-transform based, were compared on the problem of finding $\RSTF$ by simulating a force-calibration scenario where a random $\RSTF$ and 100 random poses $\RRBTF$ were generated. In one simulation, we let the first algorithm find $\RSTF$ and $m$ with an error in the initial estimate of $m$ by a factor of 2, while the Cayley algorithm was given the correct mass. The results, depicted in the left panel of~\cref{fig:calibforce} indicate that the two methods performed on par with each other. The figure shows the error in the estimated rotation matrix as a function of the added measurement noise in $f_S$. In a second experiment, depicted in the right panel of~\cref{fig:calibforce}, we started both algorithms with a mass estimate with \SI{10}{\%} error. Consequently, the Cayley algorithm performed significantly worse for low noise levels, while the difference was negligible when the measurement noise was large enough to dominate the final result.


The experiment showed that not only does the relaxation-based method perform on par with the unconstrained Cayley-transform based method, it also allows us to estimate the mass, reducing the room for potential errors due to an error in the mass estimate.
It is thus safe to conclude that the relaxation-based algorithm is superior to the Cayley algorithm in all situations. Implementations of both algorithms are provided in~\softwarecite{Robotlib}.

We provide no numerical evaluation of finding the the vector to the center of gravity, $r$, as this is a standard linear least-squares problem once the gravity vector and the rotation matrix are estimated.

\subsection{Unknown gravity vector}
\begin{figure}[tpb]
    \centering
    \setlength{\figurewidth}{0.34\linewidth}
    \setlength{\figureheight }{4.2cm}
    \input{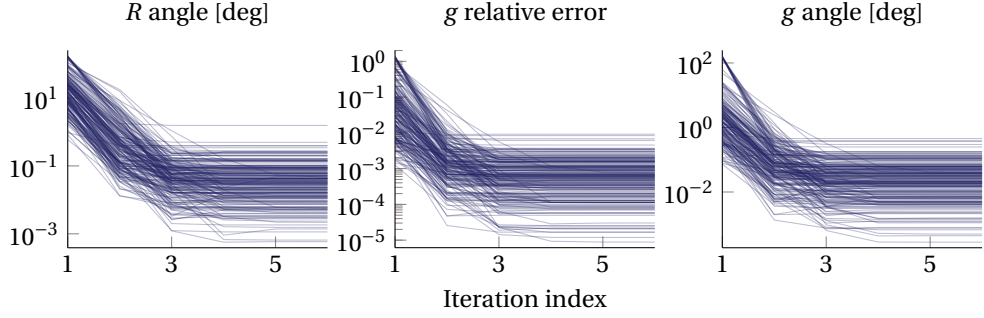}
    \caption{Errors vs. iteration index during joint estimation of rotation matrix and gravity vector. $N=100$ points were randomly sampled using a gravity vector sampled randomly. Measurement noise of $\sigma_f = \SI{1}{Nm}$ was added for a signal-to-noise ratio of $\sigma_g/\sigma_f = 100$. The procedure was repeated 200 times.}
    \label{fig:gravityerrors}
\end{figure}

We evaluated the performance of the nullspace method, the eigendecomposition-based method and the iterative least-squares method by randomly sampling data like above, and computing errors vs. iteration index of the loop in \cref{alg:iterative}.
\Cref{fig:gravityerrors} displays results vs. iteration for the iterative method. The errors are the angle between $\hat{R}_S^{\scriptscriptstyle TF}$ and $\RSTF$, calculated as $\cos^{-1}\big( (\operatorname{trace}(R_1\T R_2)-1)/2$\big), the relative error in $\hat g$ calculated as $\norm{\hat g - g}/\norm{g}$ as well as the error in the direction of $g$ calculated as
$$\cos^{-1}\Big(\dfrac{g\T \hat g}{ \norm{g}\norm{\hat g}}\Big)$$
The initial guess for the gravity vector, $g_0$, was a random vector of unit variance, while the true gravity vector, $g$, was a random vector with variance $\sigma_g^2 = 100^2$, corresponding to a random direction and a random mass with standard deviation of approximately \SI{7}{\kg}. The procedure was repeated 200 times with new random samples.

\Cref{fig:gravityerrors} indicates that the procedure converged in all cases, even though the initial guess $g_0$ was wrong by several orders of magnitude, including cases where $g\T g_0 < 0$.
The figure illustrates how the relative error in $\hat g$ typically converged to $<1\%$. This is of course dependent on the signal-to-noise ratio and scales as expected.

The nullspace method and the eigen-decomposition method find exactly the same solution as the iterative method. This indicates that the iterative method can be seen as a power iteration that converges to the largest eigenvector of $\mathcal{K}$.

\section{Discussion and Conclusions}
This work has provided methods that solve for the unknown rotation matrix, and the gravitational acceleration vector (magnitude and direction) relating the measurements of a force sensor to the frame where it is mounted. We also provided a method for estimating the vector to the center of gravity of the mass held by the sensor for calibration of torque measurements.  No special equipment was required for any of the methods.

The fact that all presented methods find the exact same solution indicates that they are in some sense equivalent. 

We found that of practical importance for the noise sensitivity of the nullspace method is to form all pairwise differences of measurements.

\bibliography{mypapers.bib,software.bib,merged.bib,forcecalib_arxiv.bib}
\bibliographystyle{plain}
\appendix
\section{Projection onto $SO(3)$}
\label{sec:orthonormal}

A matrix $R$ is said to be orthonormal if $R\T R = RR\T = I$. If the additional fact $\det(R) = 1$ holds, the matrix is said to be a rotation matrix, an element of the $n$-dimensional special orthonormal group $SO(n)$ \cite{murray1994mathematical, mooring1991fundamentals}.

Given an arbitrary matrix $\tilde{R} \inspace{3\times 3}$, the closest rotation matrix in $SO(3)$, in the sense $||R-\tilde{R}||_F$, can be found by Singular Value Decomposition according to \cite{eggert1997estimating}
\begin{align}
    \tilde{R} &= USV\T\\
    R &= U \begin{bmatrix}1 & &\\ & 1 & \\ & & \det(UV\T) \end{bmatrix} V\T
\end{align}

\pagebreak
\section{Implementations}\label{sec:implementations}
\begin{algorithm}
\caption{Julia code to solve for $R$ and $g$ using the proposed methods.}
\label{alg:juliaeigen}
\begin{multicols}{2}
\begin{minted}[breaklines, numbersep=3pt, gobble=0, fontsize=\footnotesize, framesep=1mm]{julia1}
using LinearAlgebra, FillArrays
getDF(RRBTF, forces) =
    (reshape(RRBFT, 3, :)', # D
     kron(forces, Eye(3)))  # F

"""
    R,g = calibForceEigen(RRBFT,forces)
- `RRBFT` is a 3x3xN array of R_RB^TF
- `forces` is a Nx3 matrix of force measurements
"""
function calibForceEigen(RRBFT,forces)
    D,F = getDF(RRBTF, forces)
    K = F\D*(D\F)  # Solve
    R = eigenR(K)
    g = D\F*vec(R) # Least-squares estimate of g
    R,g
end

function project(R)
    U,S,V = svd(R)
    a = sign(det(U*V'))
    S = diagm(0=>[1,1,a])
    R = U*S*V'
end

function toR(r)
    R = reshape(r,3,3)
    det(R) < 0 && (R = -R)
    project(R) # Project onto SO(3)
end

function eigenR(K)
    v = real(eigen(K).vectors[:,1])
    toR(r)
end
\end{minted}
\end{multicols}
\begin{minted}[breaklines, numbersep=3pt, gobble=0, fontsize=\footnotesize, framesep=1mm]{julia1}
function calibForceNullspace(RRBFT,forces)
    N  = size(RRBFT,3)
    I3 = Eye(3)
    M  = [RRBFT[:,:,k]'*RRBFT[:,:,i]*kron(forces[i,:]', I3) - kron(forces[k,:]', I3) for i = 1:N-1 for k=i+1:N]
    M    = reduce(vcat,M)
    R    = toR(svd(M).V[:,end]) # Form R from the nullspace of M
    D,F3 = getDF(RRBTF, forces)
    g    = D\F3*vec(R) # Least-squares estimate of g
    R,g
end

function calibForceIterative(RRBFT,forces)
    D,F3 = getDF(RRBTF, forces)
    DF   = D\F
    K    = (F\D)*DF
    R    = Eye(3) # Initialize
    r    = vec(R)
    local g
    for iter = 1:6
        g = DD*r   # Least-squares estimate of g
        r = FF*g   # Least-squares estimate of r
        R = toR(r) # Reshape and project onto SO(3)
        r = vec(R)
    end
    R,g
end
\end{minted}
\end{algorithm}

\end{document}